\documentclass[runningheads]{llncs}
\usepackage{graphicx}

\usepackage{multirow}
\usepackage{multicol}
\usepackage{amsmath}
\usepackage{xcolor}
\usepackage{array}
\usepackage{footmisc}
\usepackage{calc}
\usepackage{subfigure}
\usepackage{booktabs}
\usepackage[colorlinks=true,linkcolor=black,citecolor=blue,urlcolor=blue]{hyperref}

\usepackage{cleveref}
\usepackage[most]{tcolorbox}
\begin{document}

\title{Reasoning Through Memorization: Nearest Neighbor Knowledge Graph Embeddings}
\newcommand{\ours}{$k$NN-KGE}
\newcommand{\repeatthanks}
{\textsuperscript{\thefootnote}}
\let\origthanks\thanks
\renewcommand\thanks[1]{\begingroup\let\rlap\relax\origthanks{#1}\endgroup}

%
%
\author{
Peng Wang$^{1}\thanks{Equal contribution.}$,
Xin Xie$^{1}\repeatthanks$,
Xiaohan Wang$^{1}$,
Ninyu Zhang$^{1\dagger}$
}


%
\titlerunning{Reasoning Through Memorization: \ours}
%

\institute{$^{1}$Zhejiang University \\
\email{\{peng2001,xx2020,wangxh07,zhangningyu\}@zju.edu.cn} \\ 
}

\maketitle              
\begin{abstract}
Previous knowledge graph embedding approaches usually map entities to representations and utilize score functions to predict the target entities, yet they typically struggle to reason rare or emerging unseen entities. In this paper, we propose \emph{k}NN-KGE, a new knowledge graph embedding approach with pre-trained language models, by linearly interpolating its entity distribution with \emph{k}-nearest neighbors. We compute the nearest neighbors based on the distance in the entity embedding space from the knowledge store. Our approach can allow rare or emerging entities to be memorized explicitly rather than implicitly in model parameters. Experimental results demonstrate that our approach can improve inductive and transductive link prediction results and yield better performance for low-resource settings with only a few triples, which might be easier to reason via explicit memory\footnote{Code is available at: \url{https://github.com/zjunlp/KNN-KG}}.


\end{abstract}
\section{Introduction}

Knowledge Graphs (KGs) organize facts in a structured way as triples in the form of \texttt{<subject, predicate, object>}, abridged as $(s, p, o)$, where $s$ and $o$ denote entities and $p$ builds relations between entities. 
Most KGs are far from complete due to emerging entities and their relations in real-world applications; hence KG completion—the problem of extending a KG with missing triples—has appeal to researchers \cite{transR,DBLP:conf/www/ZhangDSCZC20,DBLP:conf/ijcai/QiZCCXZZ21,DBLP:conf/kdd/ZhangJD0YCTHWHC21,DBLP:conf/ijcai/Chen0YCDHC22,DBLP:conf/sigir/Chen0ZZYXC22,张宁豫2022knowledge,Chen_2022}.

\begin{figure}[htbp]
    \centering
    \includegraphics[width=0.7\textwidth]{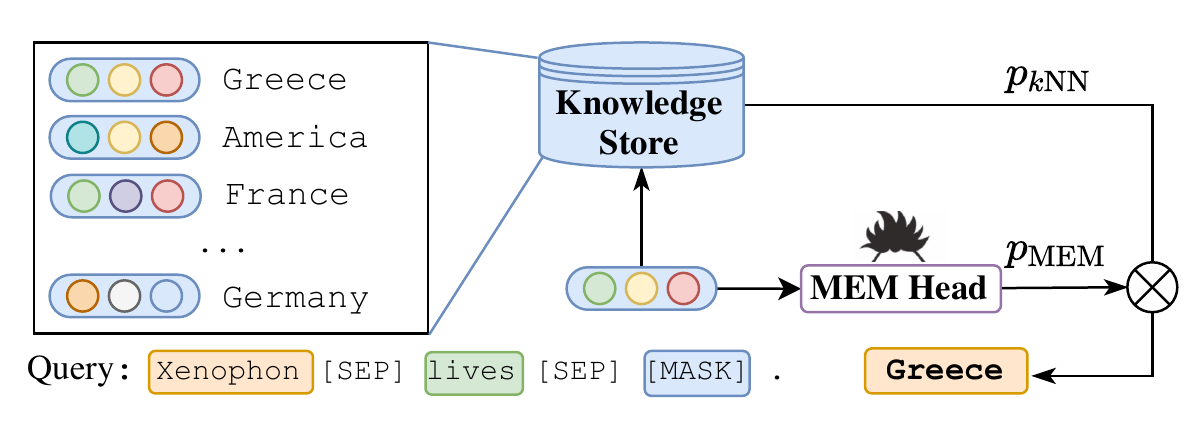}
    \caption{Our \ours~not only leverages entity prediction from softmax (MEM head in PLMs) but also retrieves the entities from the knowledge store constructed from entity descriptions and training triples.}
    \label{fig:intro}
\end{figure}
Early, traditional KG completion methods, such as TransE \cite{Bordes:TransE}, ComplEx \cite{complex}, and RotatE \cite{RotatE}, are knowledge embedding techniques that embed the entities and relations into a vector space and then obtain the predicted triples by leveraging a pre-defined scoring function to those vectors. 
Recently, another kind of KG embedding method utilize textual descriptions with language models, e.g., KG-BERT \cite{kgbert}, StAR \cite{STAR}, and KGT5 \cite{kgt5}, which are increasingly promising techniques with empirical success.
Note that those methods implicitly encode all of the relational knowledge in the weights of the parametric neural network via end-to-end training.
However, a major limitation of these previous approaches is that they can hardly reason through rare entities evolving in a few triples or emerging entities unseen during training. 
 
Note that human reasoning is facilitated by complex systems interacting together, for example, integration of current knowledge and retrieval from memory. 
Recent progress in memory-augmented neural networks has given rise to the design of modular architectures that separate computational processing and memory storage. 
Those memory-based approaches (or non/semi-parametric methods) have been applied to tasks such as language modeling \cite{DBLP:conf/iclr/KhandelwalLJZL20} and question answering \cite{DBLP:conf/emnlp/KassnerS20}, which are expressive and adaptable. 

Inspired by this, we propose \ours, an approach that extends knowledge graph embedding with language models by linearly interpolating its entity distribution with a $k$-nearest neighbors ($k$NN) model. 
As shown in Fig \ref{fig:intro}, we construct a knowledge store of entities with pre-trained language models (PLMs) and retrieve nearest neighbors according to distance in the entity embedding space. 
Given a triple with a head or tail entity missing, we utilize the representation of \texttt{[MASK]} output as the predicted anchor entity embedding to find the nearest neighbor in the knowledge store and interpolate the nearest-neighbor distribution with the masked entity prediction. 
Thus, rare entities or emerging triples can be memorized explicitly, which makes reasoning through memorization rather than implicitly in model parameters.
Experimental results on two datasets (FB15k-237 \cite{fb15k}, and WN18RR \cite{wn18rr}) in both transductive and inductive reasoning demonstrate the effectiveness of our approach. 
We further conduct a comprehensive empirical analysis to investigate the internal mechanism of {\ours}.
Qualitatively, we observe that our approach is particularly beneficial for low-resource knowledge graph embedding, which might be easier to access via explicit memory.
 Our contributions can be summarized as follows:
\begin{itemize}
    \item To the best of our knowledge, this is the first semi-parametric approach for knowledge graph embedding. 
    Our work may open up new avenues for improving knowledge graph reasoning through explicit memory.
    
    \item We introduce \ours~that can explicitly memorize rare or emerging entities, which is essential in practice since KGs are evolving. 
    
    \item Experimental results on two benchmark datasets with transductive and inductive settings show that our model can yield better performance than baselines and is particularly beneficial for low-resource reasoning.
\end{itemize}

\section{Methodology}

\begin{figure*}[!htbp]
\centering 
\includegraphics[width=1.\textwidth]{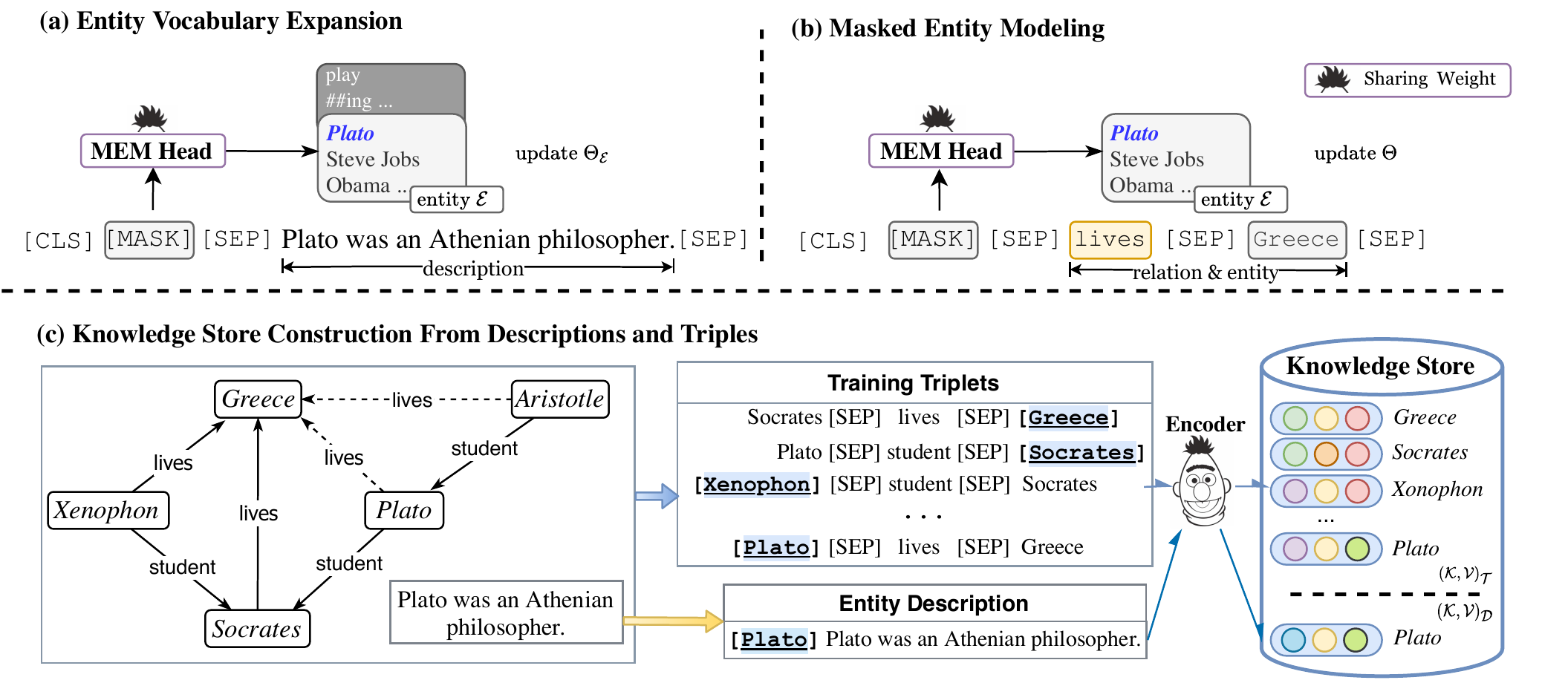} 
\caption{
The virtual entity token embedding $\Theta_{\mathcal{E}}$ in the word embedding layer (head) is firstly optimized by Entity Vocabulary Expansion (Fig a).
Then the model $\Theta$ shares the weight $\Theta_{\mathcal{E}}$ in Masked Entity Modeling (Fig b) for training.
Finally, with the model trained above, the entities in triples and descriptions colored in \textcolor{blue}{blue} will be encoded into contextualized entity representation and added to our knowledge store (Fig c).
Examples are taken from \protect\cite{liu2021indigo}.
}
\label{fig:model}
\end{figure*}

\subsection{Preliminary}

\textbf{Knowledge Graph}.
We define a knowledge graph with entity descriptions as a tuple $\mathcal{G}=(\mathcal{E}, \mathcal{R}, \mathcal{T}, \mathcal{D})$, where $\mathcal{E}$ represents a set of entities, $\mathcal{R}$ represents relation types, $\mathcal{T}$ represents a set of triples and $ \mathcal{D}$ represents the entity descriptions. 
For each triple in $\mathcal{T}$, it has the form $(e_i, r_j, e_k)$ where $e_i, e_k \in \mathcal{E}$ is the head and tail entity respectively. For each entity $e_i \in \mathcal{E}$, there exists a text $d_i$ to describe $e_i$.
To complete missing triples in knowledge graphs, link prediction is proposed, which aims at predicting the tail entity given the head entity and the query relation, denoted by $(e_i, r, ?)$\footnote{or head entity prediction denoted by $(?, r, e_i)$},  where the answer is supposed to be always within the KG.


\textbf{MEM Head}.
Like the ``word embedding layer'' (MLM Head) in the pre-trained language model that maps contextualized token representation into probability distribution of tokens in the vocabulary, MEM Head which consists of entity embeddings, maps contextualized entity representation to the probability distribution of the entity in the knowledge graph.
We will illustrate the details of using MEM head in Section \ref{sec:memorized inference}.

\subsection{Framework}

In this Section, we introduce the general framework of the proposed approach as shown in Fig \ref{fig:model}. 
We first propose masked entity modeling and entity vocabulary expansion in Section \ref{sec:contextualized kg representation}, which converts link prediction into an entity prediction task.
To address the issue of rare or unknown entities,  in Section \ref{sec:knowledge store}, we construct a knowledge store based on entity descriptions and triples in the training set to retrieve the entity by the anchor embedding (\textbf{the representation of \texttt{[MASK]} output}) during training.
Lastly, in Section \ref{sec:memorized inference}, we provide the details of inference, which makes reasoning through memorization rather than implicitly in model parameters.

\subsection{Contextualized KG Representation}
\label{sec:contextualized kg representation}

\label{sec:masked entity prediction}

In this subsection, we treat the BERT model as the entity predictor because we convert the link prediction task to a masked entity modeling task, which uses the structural information or text description to predict the missing entity.

\textbf{Masked Entity Modeling}.
For link prediction, given an incomplete triple $(e_i,r_j,?)$, previous studies utilize KG embeddings or textual encoding to represent triple and leverage a pre-defined scoring function to those vectors. 
In this paper, we simply leverage masked entity modeling for link prediction, which makes the model predict the correct entity $e_k$ like the Masked Language Model (MLM) task.
The model only needs to predict the missing entity at the tail or head\cite{zhang2023multimodal}.

Specifically, given a querying triple $(e_i, r_j, ?)$ and the description $d$ to the entity $e_i$, we concatenate this triple and the entity description $d$ to obtain the input sequence $x_k$ to predict the entity $e_k$ as follows:
$$
x_k = \texttt{[CLS]} \ e_i \ d \ \texttt{[SEP]} \ r_j  \ \texttt{[SEP]} \ \texttt{[MASK]} \ \texttt{[SEP]}.
$$

By masked entity modeling, the model can obtain the correct entity $e_k$ by ranking the probability of each entity in the knowledge graph with $p_{\text{MEM}}$.
\begin{equation}
    p_{\text{MEM}}(y|x) = p(\texttt{[MASK]} = e_k | x_k ; \Theta),
\label{eq:p_mem}
\end{equation}
where $\Theta$ represents the parameters of the pre-trained language models.

\let\mc\multicolumn

\begin{table*}[thbp]
\caption{Inference efficiency comparison. $|d|$ is the length of the entity description. $|\mathcal{E}|$, $|\mathcal{R}|$ and $|\mathcal{T}|$ are the numbers of all unique entities, relations and triples in the graph respectively. Usually, $|\mathcal{E}|$ exceeds hundreds of thousands and is much greater than $|\mathcal{R}|$.}
\label{tab:results-inductive}
\centering
\resizebox{1.0\textwidth}{!}{
\begin{tabular}{lcccc}
\toprule
Inference & Method & Complexity & Speed up & GPU time under RTX 3090\\
\midrule 
\multirow{2}{*}{ One Triple } & StAR    &  $O\left(|d|^{2}(1+|\mathcal{E}|)\right)$ & $\sim |\mathcal{E}| \times$ & -\\
                              & \ours   & $O\left(|d|^{2} + |\mathcal{E}| + |\mathcal{T}| \right)$ & & -\\
\midrule 
\multirow{2}{*}{ Entire Graph } & StAR  & $O\left(|d|^{2}|\mathcal{E}|(1+|\mathcal{R}|)\right)$ & $\sim 2 \times$ because $|\mathcal{E}||\mathcal{R}| \gg |\mathcal{T}|$ & 28 min\\
    & \ours & $O\left((|d|^{2}+|\mathcal{T}|+|\mathcal{E}|) |\mathcal{T}| \right)$ & & 15 min\\
\bottomrule
\end{tabular}
}%

\label{tb:time}
\end{table*}

Note that the procedure of masked entity modeling is simple yet effective, and the one-pass inference speed is faster than the previous BERT-based model like StAR \cite{STAR}.
A detailed comparison of inference time can be found in Table \ref{tb:time}.
To do so, we use the same loss function to optimize our masked entity models.
\begin{equation}
\mathcal{L_{\text{MEM}}}=-\frac{1}{\left| \mathcal{E} \right|} \sum_{e \in \mathcal{E}} 1_{e=e_k} \log p\left(\texttt{[MASK]}=e \mid x_k ; \Theta \right),
\end{equation}
where $|\mathcal{E}|$ is the number of total entities $\mathcal{E}$ and $\Theta$ represents the parameters of the model.

We propose entity vocabulary expansion to utilize embeddings for each unique entity.

\textbf{Entity Vocabulary Expansion}.
\label{sec:semantic-aware entity modeling}
Since it is non-trivial to utilize subwords for entity inference, we directly utilize embeddings for each unique entity as common knowledge embedding methods \cite{Bordes:TransE} do.
We represent the entities $e \in \mathcal{E}$ as \textbf{special tokens in language model's vocabulary}; thus, knowledge graph reasoning can reformulate as a masked entity prediction task as shown in Fig \ref{fig:intro} and Fig \ref{fig:model}.


$$ x_d = \texttt{[CLS]} \ \text{prompt}(\texttt{[MASK]}) \ \texttt{[SEP]} \ d \ \texttt{[SEP]}$$

We optimize those entity embeddings (random initialization) by predicting the entity $e_i$ at the masked position with the other parameters fixed. 
Formally, we have the following:
\begin{equation}
\mathcal{L}=-\frac{1}{\left| \mathcal{E} \right|} \sum_{e \in \mathcal{E}} 1_{e=e_i} \log p\left(\texttt{[MASK]}=e \mid x_d ; \Theta \right)
\label{loss}
\end{equation}
where $|\mathcal{E}|$ is the number of total entities $\mathcal{E}$ and $\Theta$ represents the parameters of the model.

\subsection{Knowledge Store}\label{sec:knowledge store}

Inspired by recent progress in memory-augmented neural networks \cite{DBLP:conf/iclr/KhandelwalLJZL20,DBLP:conf/emnlp/KassnerS20}, we construct a knowledge store to explicitly memorize entities. 
We introduce the details  of construction as follows:


\textbf{From Descriptions $\mathcal{D}$}.
Let $f(\cdot)$ be the function that maps the entity $e$ in the input $x$ to a fixed-length vector representation computed by PLM.
We use pre-designed prompts to obtain entity embedding from entity descriptions.
We can construct the knowledge store $(\mathcal{K}, \mathcal{V})_{\mathcal{D}}$ from descriptions (the set of descriptions $\mathcal{D}$ of all entities $\mathcal{E}$ in $\mathcal{G}$).
\begin{equation}
    (\mathcal{K}, \mathcal{V})_{\mathcal{D}}=\left\{\left(f\left(x_d\right), e_{i}\right) \mid\left(d, e_{i}\right) \in \mathcal{G}\right\}
\end{equation}

\textbf{From Triples $\mathcal{T}$}.
Since different relations focus on different aspects of the same entity, it is intuitive to utilize different triples to represent entities. 
For example, given the triple (\texttt{Plato}, \texttt{lives}, ?), the model can reason through the triple of (\texttt{Plato}, \texttt{nationality}, \texttt{Greece}) in KGs to obtain \texttt{Greece}.
Thus, we also construct a knowledge store from triples. 
We can construct the knowledge store $(\mathcal{K}, \mathcal{V})_{\mathcal{T}}$ from triples (the set of all triples $\mathcal{T}$ in $\mathcal{G}$).
\begin{equation}
    (\mathcal{K}, \mathcal{V})_{\mathcal{T}}=\left\{\left(f\left(x_t\right), e_{i}\right) \mid\left(t, e_{i}\right) \in \mathcal{G}\right\}
\end{equation}




\subsection{Memorized  Inference}\label{sec:memorized inference}

In this Section, we will introduce memorized inference, the procedure of using KNN to choose the nearest embedding in the knowledge store. 

Specifically, given a triple with the head or tail entity missing, we use the representation of \texttt{[MASK]} output as the predicted anchor entity embedding to find the nearest neighbor in the knowledge store. 
\begin{equation}
    \text{d}(\mathbf{h}_i, \mathbf{h}_j) = |\mathbf{h}_i, \mathbf{h}_j|_{\text{2}},
\end{equation}
where $||_{\text{2}}$ refers to the Euclidean distance.
$\mathbf{h}_{\texttt{[MASK]}}$ is the representation of the \texttt{[MASK]} token in the input sequence.
$\mathbf{h_i}$ and $\mathbf{h_j}$ refers to the representation of two different embeddings (here, we refer to the $\mathbf{h}_{\texttt{[MASK]}}$ and the embedding in the knowledge store, respectively).
Thus, we can obtain the probability distribution over neighbors based on a softmax of $k-\text{nearest}$ neighbors.

For each entity retrieved from the knowledge store, we choose only one nearest embedding in the knowledge store to represent the entity.
\begin{equation}
p_{\mathrm{kNN}}(y \mid x) \propto \sum_{\left(k_{i}, v_{i}\right) \in \mathcal{N}} 1_{y=v_{i}} \exp \left(-\text{d}\left(k_{i}, f(x)\right)\right)
\end{equation}
where $x$ refers to the input sequence, and $y$ refers to the target entity distribution by KNN via retrieving from the knowledge store.
We interpolate the nearest neighbor distribution $p_{k\text{NN}}$ with the model entity prediction $p_{\text{MEM}}$ which can be seen in \cref{eq:p_mem} and using a hyper-parameter  $\lambda$ to produce the final \ours~distribution:
\begin{equation}
p(y \mid x)=\lambda p_{k\mathrm{NN}}(y \mid x)+(1-\lambda) p_{\text{MEM}}(y \mid x)
\end{equation}


\section{Experiments}

We conduct extensive experiments to evaluate the performance of {\ours} by answering the following research questions:

\begin{itemize}
    \item \textbf{RQ1}: How does our {\ours} perform when competing with different types of transductive knowledge graph completion methods?
    \item \textbf{RQ2}: What is the benefits of our {\ours} when comparing with different inductive knowledge graph completion approaches?
    \item \textbf{RQ3}: How effective is our  {\ours} in reasoning with less training or long-tailed triples?
    \item \textbf{RQ4}: How do different key modules in our  {\ours} contribute to the overall performance?
    
\end{itemize}

\subsection{Experimental Settings}
We evaluate our method on FB15k-237 \cite{fb15k} and WN18RR \cite{wn18rr} in both transductive and inductive settings, which are widely used in the link prediction literature.

\begin{table*}[t]
\caption{Transductive link prediction results on WN18RR and FB15k-237. 
\ours~(w/o KS) represents we only use masked entity modeling without knowledge store.
The \textbf{bold} numbers denote the best results in each genre while the \underline{underlined} ones are the second-best performance.}
\centering
\setlength{\tabcolsep}{3.1mm}{
\resizebox{0.98\textwidth}{!}{
\begin{tabular}{lcccccccccc}
\toprule
              &         \mc{5}{c}{\bf WN18RR}                           &        \mc{5}{c}{\bf FB15k-237}                           \\ 
                        \cmidrule(lr){2-6}                         \cmidrule(lr){7-11}              
Method        &    Hits@1   &     Hits@3  &     Hits@10  &  MR       &  MRR      &    Hits@1 &   Hits@3  &  Hits@10  &       MR  &    MRR    \\ 
\midrule      
\multicolumn{11}{c}{\textit{Graph embedding approach}}                                                                              \\
\midrule
TransE \cite{Bordes:TransE}	$\diamond$    &   0.043	&   0.441	&   0.532	&   2300	&   0.243	&   0.198	&   0.376	&   0.441	&   323	    &   0.279   \\
DistMult \cite{distmult} $\diamond$	&   0.412	&   0.470	&   0.504	&   7000	&   0.444   &   0.199   &   0.301	&   0.446	&   512	    &   0.281   \\
R-GCN \cite{DBLP:conf/esws/SchlichtkrullKB18} & 0.080 &0.137 &0.207 &6700 &0.123 &0.100 &0.181 & 0.300& 600 &0.164 \\
ComplEx	\cite{complex} $\diamond$    &   0.409	&   0.469	&   0.530	&   7882	&   0.449	&   0.194	&   0.297	&   0.450	&   546	    &   0.278   \\
RotatE \cite{RotatE}	    &   0.428	&   0.492	&   0.571	&   3340	&   0.476	&   0.241	&   0.375	&   0.533	&   177	    &   0.338   \\
TuckER \cite{tucker}	    &   0.443	&   0.482	&   0.526	&   -	    &   0.470	&   0.226	&   \underline{0.394}	&   0.544	&   -	    &   \underline{0.358}   \\
ATTH \cite{ATTH}	    &\underline{0.443}	&   \underline{0.499}	&   0.573	&   -	    &   \underline{0.486}	&   \underline{0.252}	&   0.384	&   \underline{0.549}	&   -	    &   0.348   \\
\midrule
\multicolumn{11}{c}{\textit{Textual encoding approach}}                                                                             \\
\midrule
KG-BERT	\cite{kgbert}    &   0.041	&   0.302	&   0.524	&   \underline{97}	    &   0.216	&   -	    &   -	    &   0.420	&   \underline{153}	    &   -       \\
MTL-KGC \cite{kim-etal-2020-multi} & 0.203 & 0.383 & 0.597 & - & 0.331 &  0.172 & 0.298 & 0.458 & - & 0.267 \\
StAR \cite{STAR}	    &   0.243	&   0.491	&\bf0.709	&\bf51	    &   0.401	&   0.205	&   0.322	&   0.482	&\bf117	    &   0.296   \\
KGT5 \cite{kgt5} & 0.487 &  - & 0.544 & -&  0.508 & 0.210 & - & 0.414 & - & 0.276 \\
\midrule
\ours~(w/o KS) & 0.399 & 0.527 & 0.633& 1365&0.481 &0.269 & 0.396& 0.547 & 283 & 0.360 \\
\ours	    &\bf  0.525	&\bf0.604	&   \underline{0.683}	&   986	    &\bf 0.579	&\bf0.280	&\bf0.404	&\bf0.550	&   185	    &\bf0.370   \\
\bottomrule
\end{tabular}
}%
}
\label{tb:transductive}%
\end{table*}

\subsection{Transductive Experiments (RQ1)}
\label{sec:experiment_normal}



As shown in Table~\ref{tb:transductive}, our proposed \ours~can obtain state-of-the-art or competitive performance on all these datasets, especially significant improvement (0.443 $\rightarrow$ 0.525 on WN18RR and 0.252 $\rightarrow$ 0.280 on FB15k-237) in terms of Hits@1.
We hold that the improvement is mainly attributed to better retrieval by our masked entity modeling and knowledge store.

\begin{table*}[htbp]
\caption{Inductive link prediction results on WN18RR and FB15k-237 in inductive setting. $\dagger$Resulting numbers are reported by \protect\cite{inductive}.  
The \textbf{bold} numbers denote the best results in each genre while the \underline{underlined} ones are the second-best performance.
}
\centering
\resizebox{0.76\textwidth}{!}{
\begin{tabular}{lcccccccc}
\toprule
              &         \mc{4}{c}{\textbf{WN18RR} ind}                 &        \mc{4}{c}{\textbf{FB15k-237} ind}                 \\ 
                        \cmidrule(lr){2-5}                         \cmidrule(lr){6-9}              
Method        &    Hits@1   &     Hits@3  &     Hits@10  &    MRR &    Hits@1   &     Hits@3  &     Hits@10  &     MRR  \\ 
\midrule                                                                                                                                          
BE-BOW$\dagger$        & 0.045 & 0.244 & 0.450 & 0.180 & 0.103 & 0.184 & \underline{0.316} & 0.173 \\ 
BE-DKRL$\dagger$       & 0.031 & 0.141 & 0.282 & 0.139 & 0.084 & 0.151 & 0.263 & 0.144 \\
BLP-TransE$\dagger$    & 0.135 & \bf 0.361 & \bf 0.580 & \underline{0.285} & \underline{0.113} & \underline{0.213} & \textbf{0.363} & \underline{0.195} \\ 
BLP-DistMult$\dagger$  & 0.135 & 0.288 & \underline{0.481} & 0.248 & 0.076 & 0.156 & 0.286 & 0.146\\ 
BLP-ComplEx$\dagger$   & \underline{0.156} & 0.297 & 0.472 & 0.261 & 0.081 & 0.154 & 0.283 & 0.148\\ 
\midrule
\ours          & \textbf{0.223} & \underline{0.320} & 0.431 & \textbf{0.294} & \textbf{0.146} & \textbf{0.214} & 0.293 & \textbf{0.198} \\
\bottomrule
\end{tabular}
}

\label{tb:inductive}%
\end{table*}

\subsection{Inductive Experiments (RQ2)}
\label{sec:exp-inductive}

 
From Table \ref{tb:inductive}, we observe that our \ours~method can yield better or comparable performance compared with the previous state-of-the-art method, which demonstrates the effectiveness of the knowledge store.
We notice that \ours~reaches the best Hits@1, which further validates the advantages of reasoning through memorization.

\subsection{Results with  Less Training \& Long-tailed Triples (RQ3)}
\label{sec:ana-long-tail}

\begin{figure}[!htb] 
\centering 
\includegraphics[width=0.4\textwidth]{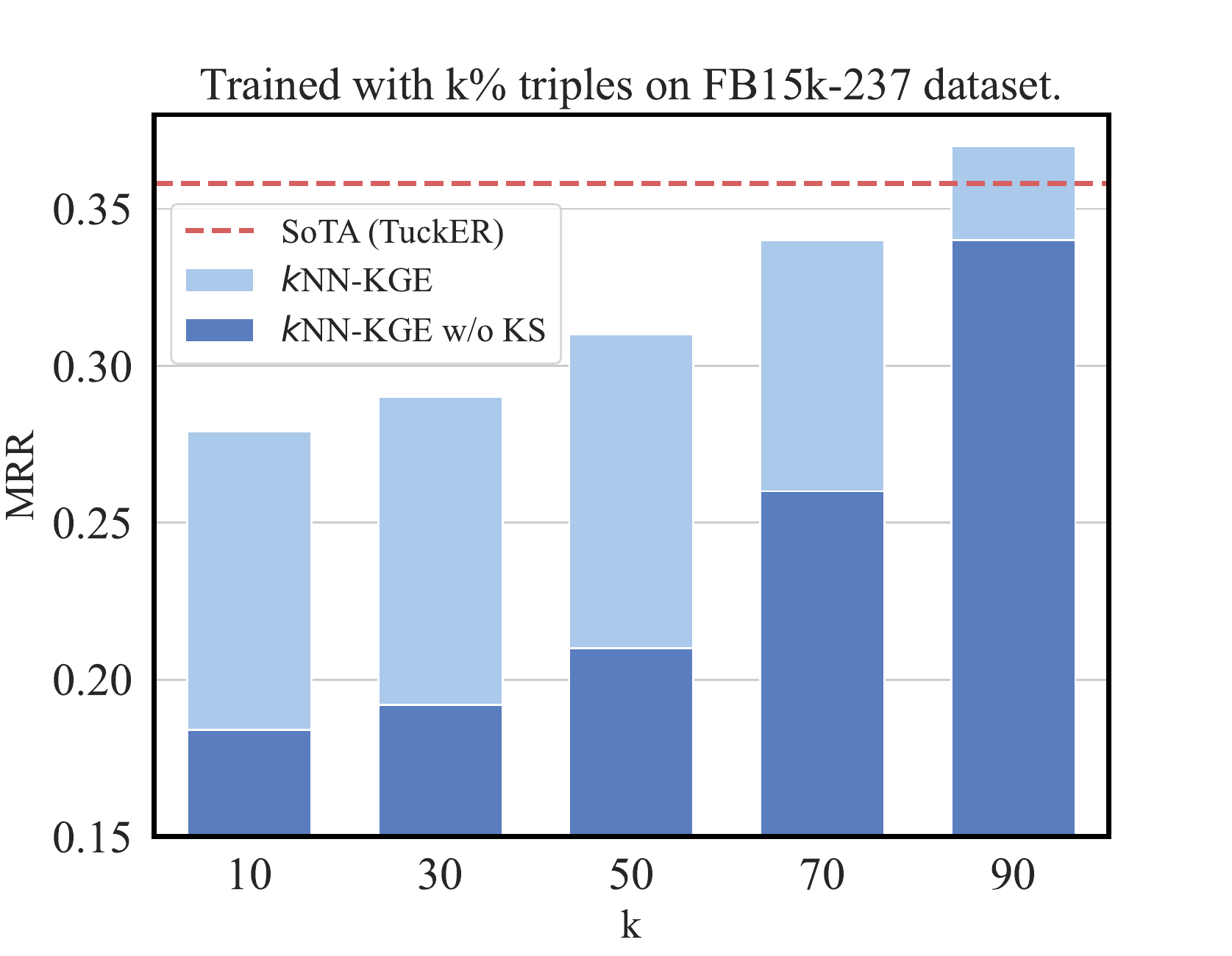} 
\caption{Varying the size of the training samples. In the low resource setting, knowledge store monotonically improves performance.} 
\label{fig:incremental}
\end{figure}
As shown in Section \ref{sec:experiment_normal} and \ref{sec:exp-inductive}, retrieving neighbors from the knowledge store can improve the model's performance.
When facing emerging entities, the translation-based model (such as TransE) have to add new entities or new triples to the models and retrain the entire model, which requires huge computation resource.  
We conduct an ablation study compared with the state-of-the-art model TuckER \cite{tucker} (red line) and our approach without retrieving knowledge store (w/o KS). 
Moreover, we notice that our approach with 70\% samples can yield comparable performance with the previous SoTA model in MRR.

\begin{figure}[htbp]
	\label{PAAbefore}
	\centering
	\subfigure[Long-tailed distribution in the FB15k-237. 82\% entities occur \textbf{$<$ 50 times}.]{
    \includegraphics[width=.47\textwidth]{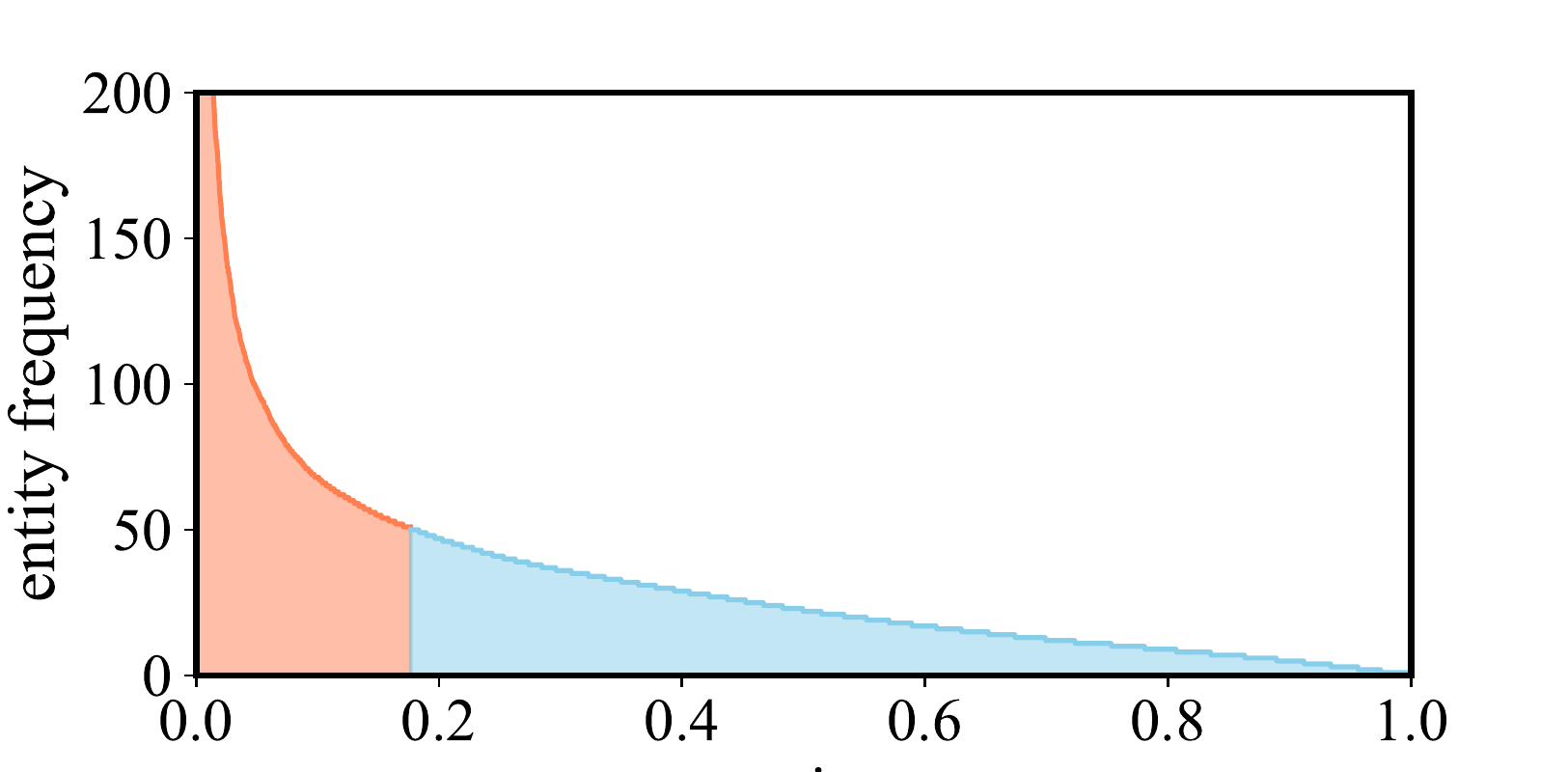} 
    }
	\subfigure[\ours~and RotatE MRR result in FB15k-237 dataset.]{

		\includegraphics[width=.47\textwidth]{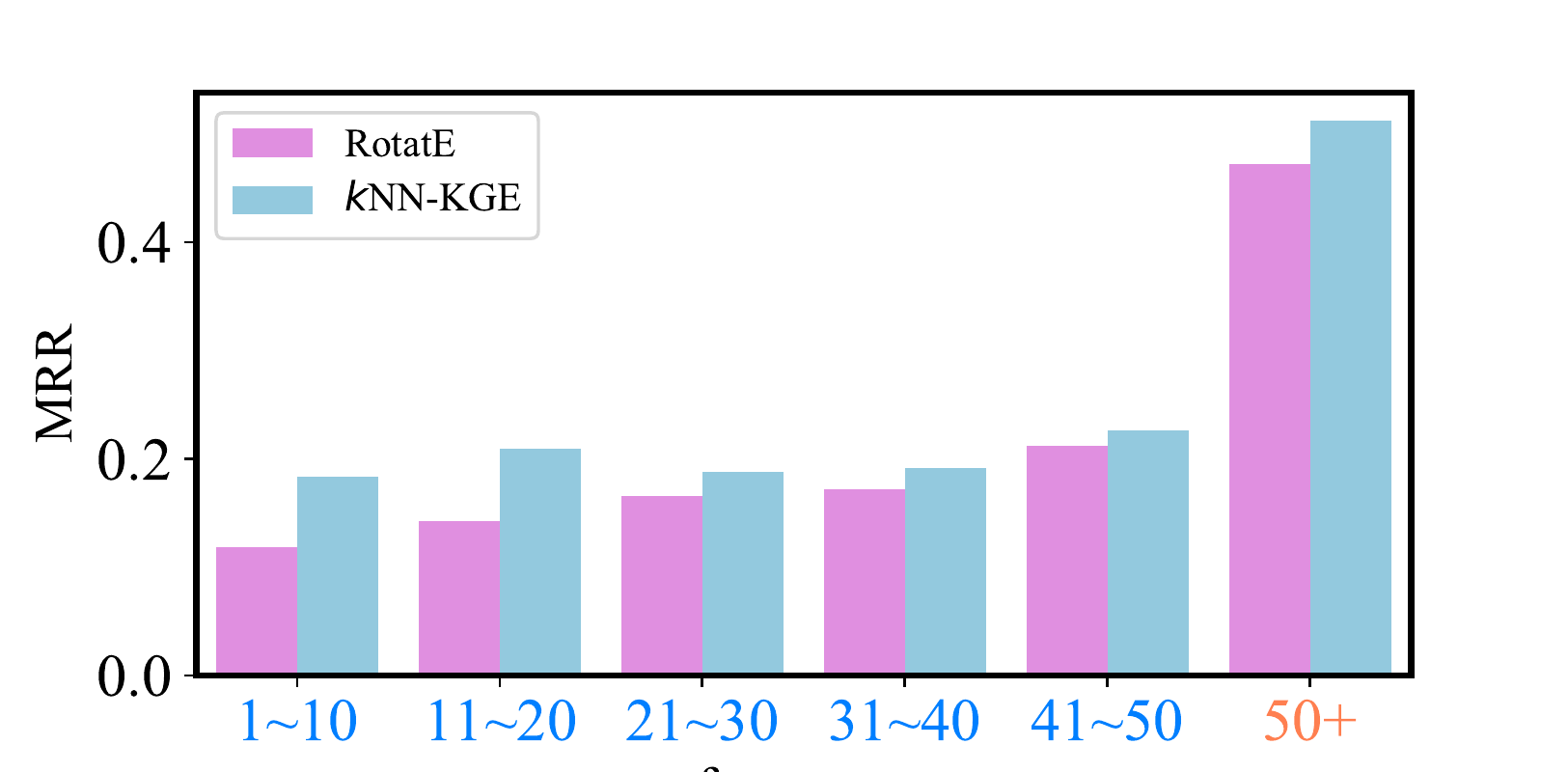} 

	}
		\caption{Significant improvement in entities with frequency below 50 colored in \textcolor{blue}{blue}.}
		\label{fig:long-tailed}
\end{figure}


\textbf{Improvements in Tail Entities.}
We find that those tail entities in long-tailed distributions can achieve performance improvement. 
We can observe from Fig \ref{fig:long-tailed} that only 18\% entities show up more than 50 times and {\ours} obtain improvements (0.118 $\rightarrow$ 0.183) with the entities occurring less than 20 times.

\subsection{Analysis (RQ4)}
\label{sec:analysis}

In this Section, we first conduct a case study and show the improvements in tail entities by the knowledge store.
Second, we visualize the entity embeddings in the knowledge store with a query triple.
Then, we analyze the effect of hyperparameters on retrieving the entities in the knowledge store.
Lastly, we summarize the size of the knowledge store and the speed of reasoning on two datasets: WN18RR and FB15k-237.

\begin{figure}[!htb] 
\centering 
\includegraphics[width=0.7\textwidth]{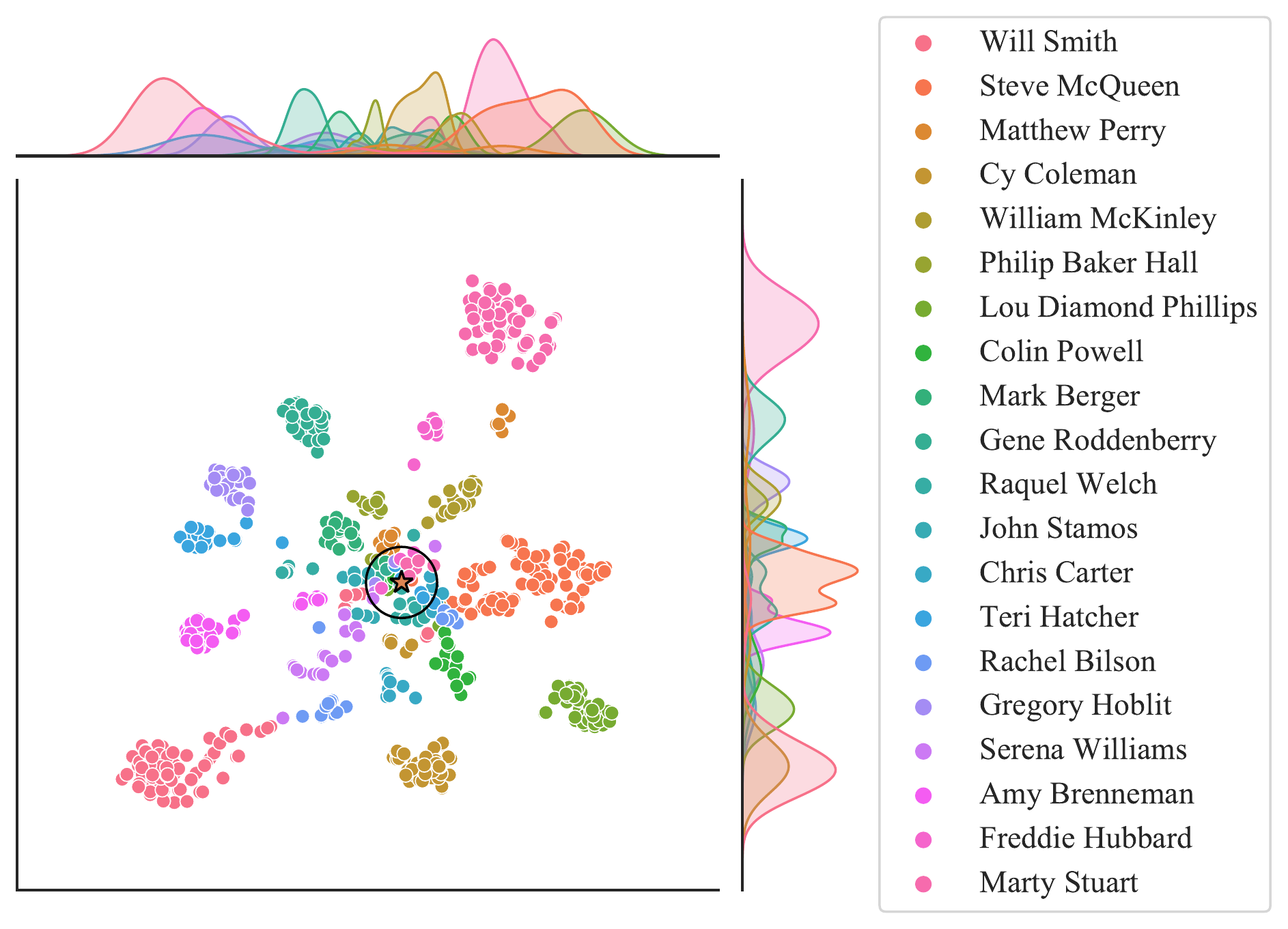} 
\caption{2D t-SNE visualisation of 20 nearest neighbor entities with all their embeddings in knowledge store retrieved by predicting the missing entity in triple (\texttt{?}, \texttt{nationality}, \texttt{America}).
Anchor embedding is marked as $\star$.} 
\label{fig:visualization}
\end{figure}

\subsubsection{\textbf{Case Study}}
We also conduct case studies to analyze the different reasoning results with or without querying the knowledge store. 
From Table \ref{tb:query diff}, we observe that \ours~w/o Knowledge Store can infer better entities given head/tail entities and relations, further demonstrating the effectiveness of the proposed approach.
Note that \ours~can explicitly memorize those entities in Knowledge Store; thus, it can directly reason through memorization.

\begin{table}[!htb]

\centering
\caption{First five entities with their probability predicted by \ours~w/o Knowledge Store, and its reranking with \ours, for two example queries.}
\resizebox{0.7\textwidth}{!}{
\begin{tabular}{cll}
\toprule
\mc{3}{l}{\textbf{Query:}\texttt{(?,ethnicity,Timothy Spall)}} \\
\midrule
Rank & \ours~w/o Knowledge Store & \ours \\
\midrule
1 & Jewish people (0.285)  & \textbf{English people} (0.548) \\
2 & \textbf{English people} (0.202) & Jewish people (0.028)\\
3 & British people (0.132) & British people (0.026)\\
4 & White British (0.080)  & Irish people in ... (0.021)\\
5 & Scottish people (0.037)& Irish people (0.018)\\
\bottomrule
\end{tabular}
}

\label{tb:query diff}
\end{table}



\subsubsection{\textbf{Visualization of Entity Embeddings in Knowledge Store}}
Since we build our knowledge store with different entity embeddings from different aspects of the same entity, we visualize(Fig \ref{fig:visualization}) those embeddings in the knowledge store.
Specifically, we random sample an input triple (\texttt{?}, \texttt{nationality}, \texttt{America}),  retrieve and visualize the 20 nearest neighbor entities with their embeddings in the knowledge store.

\begin{figure}[htbp]
    \centering%
    \begin{minipage}[b]{0.48\textwidth}
        \centering%
        \includegraphics[width=1\textwidth]{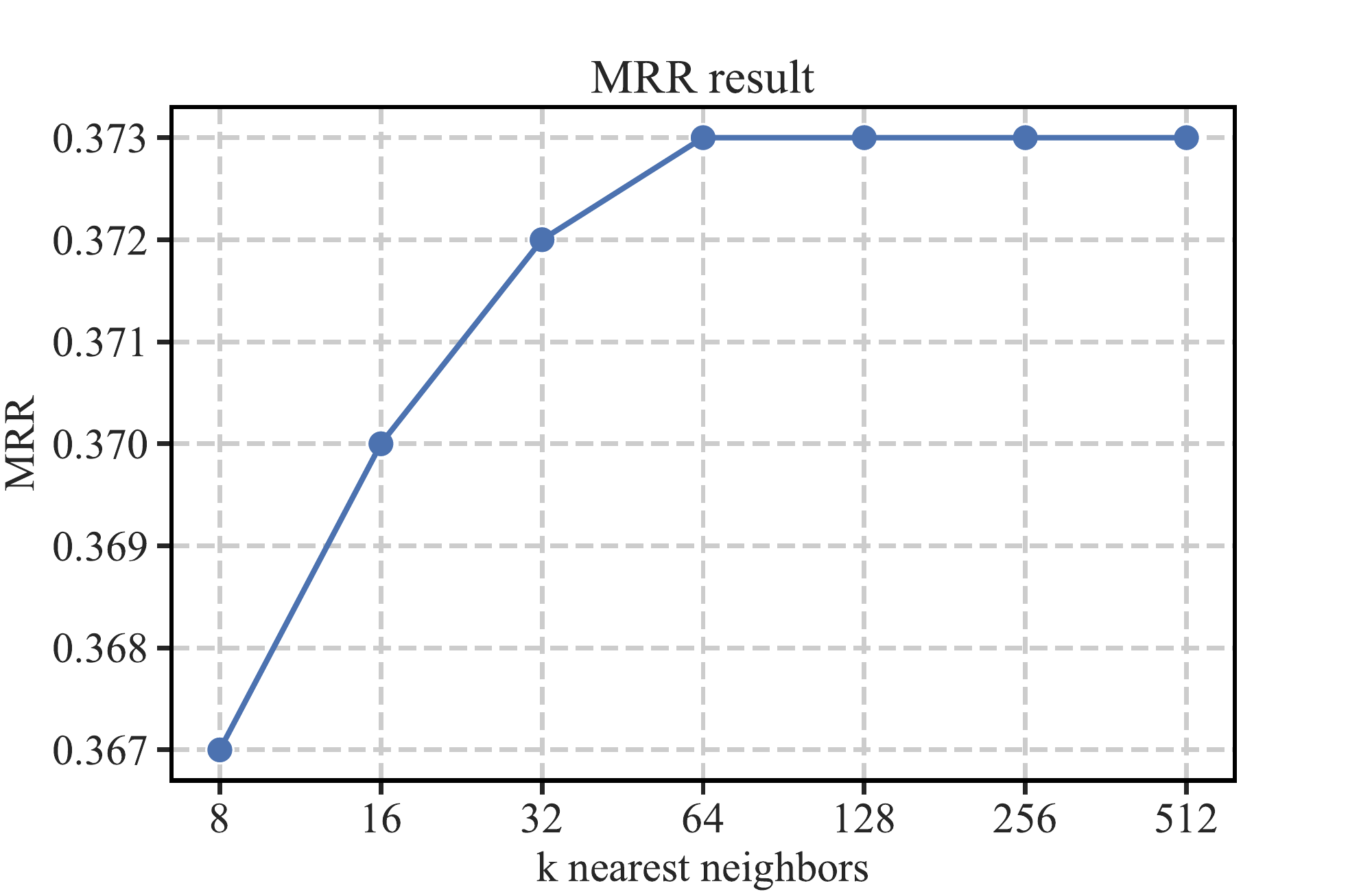}
        \caption{Effect of the number of neighbors from knowledge store.}
        \label{fig:topk}
    \end{minipage}%
\hspace{3mm}%
    \begin{minipage}[b]{0.48\textwidth}
        \centering%
        \includegraphics[width=1\textwidth]{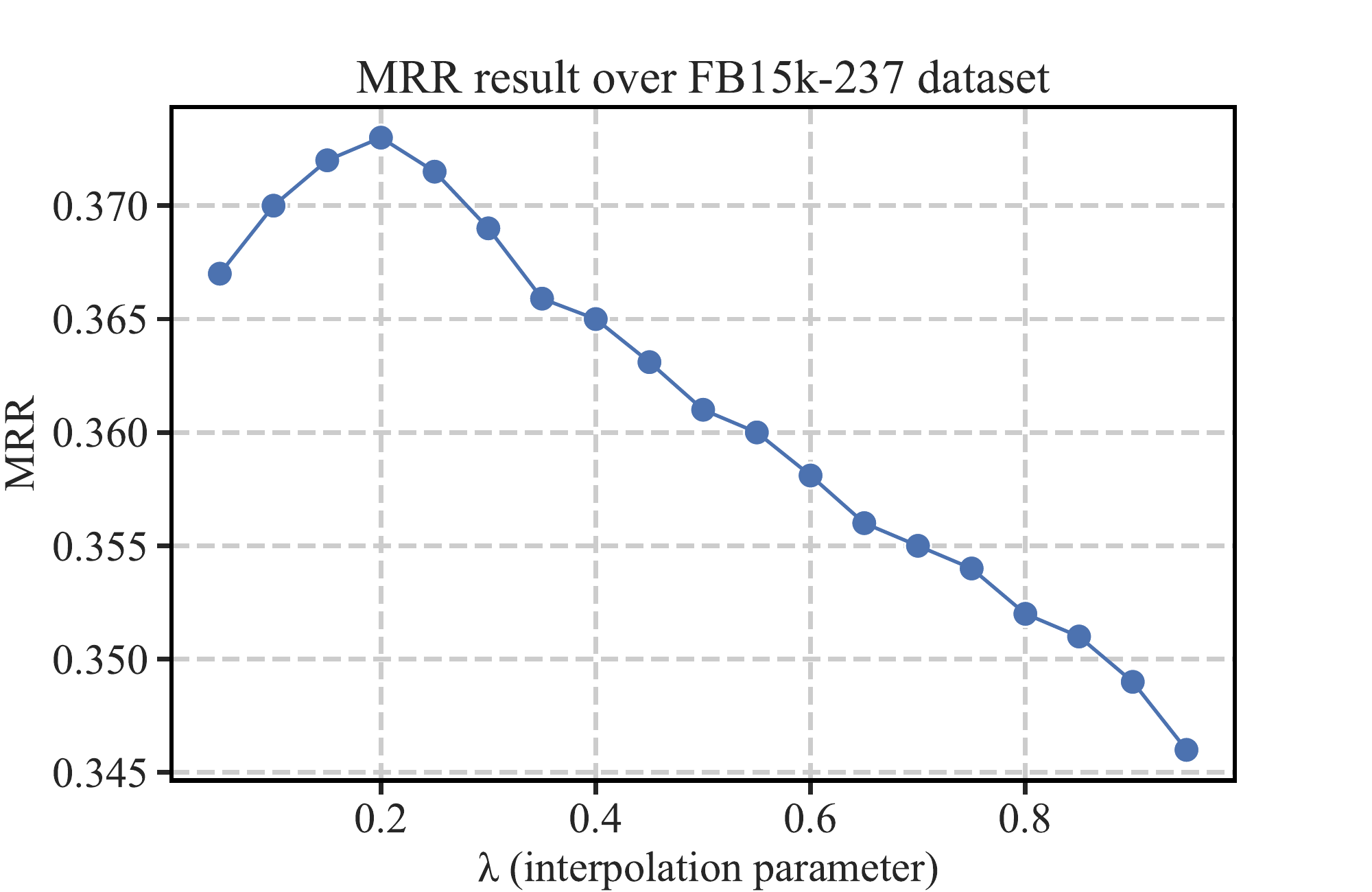}
        \caption{Effect of interpolation parameter $\lambda$ on FB15k-237 dataset.}
        \label{fig:lambda}
    \end{minipage}
\end{figure}

\subsubsection{\textbf{Impact of Number of Neighbors}}
To validate the impact of the number of neighbors, we conduct experiments with the different numbers of $k$ nearest neighbor entities to retrieve from the knowledge store.

From  Fig \ref{fig:topk}, we find that the model performance continues to improve as $K$ increases until it converges when reaching a threshold ($k=64$).
We think this is because those entities retrieved from the knowledge store are far from the anchor embedding, thus, having a low influence on the knowledge graph reasoning results.

\subsubsection{\textbf{Impact of Interpolation}}

Since we use a parameter $\lambda$ to interpolate between the BERT model distribution and the retrieved distribution from the $k$NN search over the dataset, we further conduct experiments to analyze the interpolation.

From Fig \ref{fig:lambda}, we notice that $\lambda$ = 0.2 is optimal on the FB15k-237 dataset. 
The suitable $\lambda$ can help correct the model from inferring the wrong entities, which can be seen in Table \ref{tb:query diff}.

\section{Conclusion and Future Work}

In this paper, we propose a novel semi-parametric approach for knowledge graph embedding dubbed {\ours}, which can outperform previous knowledge graph embedding models in both transductive and inductive settings by directly querying entities at test time. 
The success of \ours~suggests that explicitly memorizing entities can be helpful for knowledge graph reasoning\cite{qiao2023reasoning,zhang2023multimodal}. 

In the future, we plan to
1) improve the efficiency of the {\ours} with smaller knowledge store and faster inference speed;
2) explore how to edit\cite{yao2023editing} and delete entities from the knowledge store for dynamic KG reasoning;
3)  apply our approach to other tasks, such as question answering and fact verification.


\appendix

\bibliography{acl,custom}
\bibliographystyle{splncs04}

\end{document}